\documentclass[conference]{IEEEtran}
\IEEEoverridecommandlockouts
\usepackage{cite}
\usepackage{amsmath,amssymb,amsfonts}
\usepackage{algorithmic}
\usepackage{graphicx}
\usepackage{textcomp}
\usepackage{xcolor}
\def\BibTeX{{\rm B\kern-.05em{\sc i\kern-.025em b}\kern-.08em
    T\kern-.1667em\lower.7ex\hbox{E}\kern-.125emX}}
\usepackage{graphics} % for pdf, bitmapped graphics files
\usepackage{epsfig} % for postscript graphics files
\usepackage{mathptmx} % assumes new font selection scheme installed
\usepackage{times} % assumes new font selection scheme installed
\usepackage{amsmath} % assumes amsmath package installed
\usepackage{inputenc}
\usepackage{siunitx}
\usepackage{nicefrac}
\usepackage{xcolor}
\usepackage{eqparbox}
\usepackage{tabularx}
\usepackage{todonotes}
\usepackage{csquotes}
\usepackage{url}
\usepackage[caption=false,font=footnotesize]{subfig}
\usepackage{stmaryrd} % for some symbols like \varoast (* within a circle)
\usepackage{makecell}
\usepackage[acronym, toc, shortcuts, nowarn]{glossaries}
\glsdisablehyper
\setacronymstyle{short-long}
\newacronym{LSTM}{LSTM}{Long Short-Term Memory}
\newacronym{RMSE}{RMSE}{Root-Mean-Square Error}
\newacronym{MAP}{MAP}{Multiply-Add-Permutate}
\newacronym[plural=BSCs, longplural=Binary Spatter Codes]{BSC}{BSC}{Binary Spatter Code}
\newacronym{NGSIM}{NGSIM}{Next Generation Simulation}
\newacronym{LIDAR}{LIDAR}{Light Detection and Ranging}
\newacronym{SPA}{SPA}{Semantic Pointer Architecture}
\newacronym{Spaun}{Spaun}{Semantic Pointer Architecture Unified Network}
\newacronym{SNN}{SNN}{Spiking Neural Network}
\newacronym[plural=VSAs, longplural=Vector Symbolic Architectures]{VSA}{VSA}{Vector Symbolic Architecture}
\newacronym[plural=HRRs, longplural=Holographic Reduced Representations]{HRR}{HRR}{Holographic Reduced Representation}
\newacronym{DFT}{DFT}{Discrete Fourier Transform}
\newacronym{IDFT}{IDFT}{Inverse Discrete Fourier Transform}
\newacronym[plural=RPMs, longplural=Raven's Progressive Matrices]{RPM}{RPM}{Raven's Progressive Matrix}
\newacronym{NEF}{NEF}{Neural Engineering Framework}
\newacronym{AI}{AI}{Artificial Intelligence}

\newcommand{\norm}[1]{\left\| #1 \right\|}

\begin{document}

\title{Analyzing the Capacity of Distributed Vector Representations to Encode Spatial Information\\
}

\author{\IEEEauthorblockN{1\textsuperscript{st} Florian Mirus}
\IEEEauthorblockA{\textit{Research, New Technologies, Innovations} \\
\textit{BMW AG}\\
Garching, Germany \\
florian.mirus@bmwgroup.com}
\and
\IEEEauthorblockN{2\textsuperscript{nd} Terrence C. Stewart}
\IEEEauthorblockA{
\textit{Applied Brain Research Inc.}\\
Waterloo, Ontario, Canada \\
terry.stewart@appliedbrainresearch.com}
\and
\IEEEauthorblockN{3\textsuperscript{rd} J\"org Conradt}
\IEEEauthorblockA{\textit{Dep. of Comp. Science and Technology} \\
\textit{KTH Royal Institute of Technology}\\
Stockholm, Sweden \\
conr@kth.se}
%\and
%\and
%\and
}

\maketitle

\begin{abstract}
\aclp{VSA} belong to a family of related cognitive modeling approaches that encode symbols and structures in high-dimensional vectors. 
Similar to human subjects, whose capacity to process and store information or concepts in short-term memory is subject to numerical restrictions, the capacity of information that can be encoded in such vector representations is limited and one way of modeling the numerical restrictions to cognition.
In this paper, we analyze these limits regarding information capacity of distributed representations.
We focus our analysis on simple superposition and more complex, structured representations involving convolutive powers to encode spatial information.
In two experiments, we find upper bounds for the number of concepts that can effectively be stored in a single vector only depending on the dimensionality of the underlying vector space.
\end{abstract}


\section{Introduction}%
\label{sec:introduction}

Understanding and building cognitive systems has seen extensive research over the last decades leading to the development of several cognitive architectures.
A cognitive architecture is a \enquote{general proposal about the representation and processes that produce intelligent thought} \cite{Thagard2012}.
On the one hand, these architectures are used to explain and better understand important aspects of human behavior and intelligence.
On the other hand, they are also used to design computers and robots mimicking certain cognitive abilities of humans.

\acfp{VSA} \cite{Gayler2003} refers to a family of related cognitive modeling approaches that represent symbols and structures by mapping them to (high-dimensional) vectors. 
Such vectors are one variant of distributed representations in the sense that information is captured over all dimensions of the vector instead of one single number, which allows to encode both, symbol-like and numerical structures in a similar and unified way.
Additionally, the architectures' algebraic operations allow manipulation and combination of represented entities into structured representations.
There are several architectures such as \ac{MAP} \cite{Gayler1998}, \acp{BSC} \cite{Kanerva1988} and \acp{HRR} \cite{Plate1994}, which propose different compressed multiplication operations replacing the initially used tensor product \cite{Smolensky1990} and resulting in vectors with the same dimension as the input vectors.
One advantage of this approach is that the number of dimensions remains fixed, independent of the number of entities combined through the architecture's algebraic operations.
Schlegel et al. \cite{Schlegel2020} give an overview of eight different variants of \acp{VSA} and compare their properties and characteristics.

\begin{figure*}[t!]
    \centering
    \includegraphics[width=0.99\linewidth]{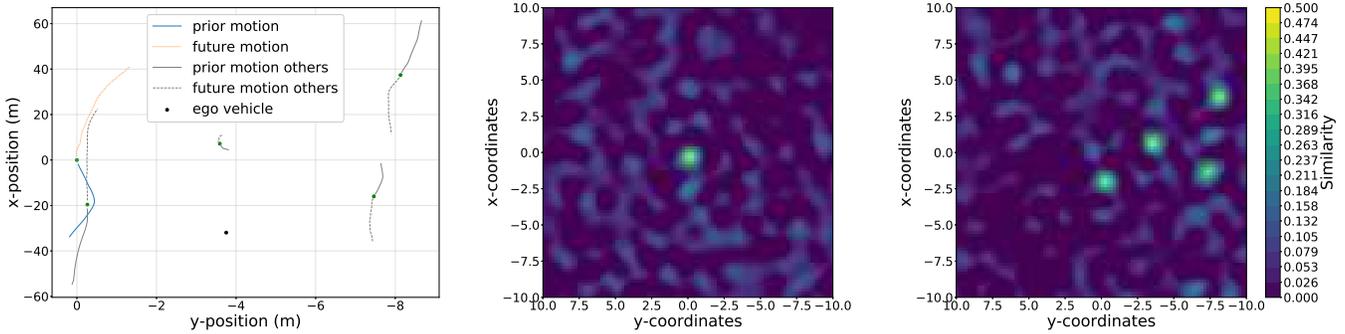}
    \caption{Visualization of the convolutive vector power encoding one particular driving scene in a \num{512}-dimensional vector.
        The left plot depicts a scene from a real-world driving data set, while the middle and right plots visualize the similarity between the representation vector of that scene and auxiliary comparison vectors created from a sequence of discrete values as heat map for the target vehicle (middle) and other cars (right).
    }
    \label{fig:spa_power_scene}
\end{figure*}

\acp{VSA} have been employed in a diverse variety of application domains, for instance, as one building block for implementing cognitive tasks such as \acp{RPM} \cite{Rasmussen2011} in \acp{SNN} \cite{Eliasmith2013} for the \ac{Spaun} model \cite{Eliasmith2012}.
Furthermore, \acp{VSA} have been used for encoding and manipulating concepts \cite{Blouw2016} as well as for human-scale knowledge representation of language vocabularies \cite{Crawford2016}.
Kleyko et al. \cite{Kleyko2015a} used \acp{VSA} to imitate the concept learning capabilities of honey bees.
In robotics \cite{Neubert2019}, \acp{VSA} have been used to learn navigation policies for simple reactive behaviors to control a Braitenberg-vehicle robot \cite{Neubert2016}.
In previous work, we proposed an automotive environment representation based on the \ac{SPA}, one particular \ac{VSA}, and employed this representation to tasks such as context classification \cite{Mirus2018} and vehicle trajectory prediction \cite{Mirus2019b}.
For the latter \cite{Mirus2019b}, we used the convolutive power of vectors to encapsulate spatial positions of several vehicles in vectors of fixed length (cf. Fig.~\ref{fig:spa_power_scene} and Sec.~\ref{sec:materials_and_methods}).
Komer et al. \cite{Komer2019} propose a similar representation of continuous space using convolutive powers and analyze it from neural perspective. 

However, given the mathematical properties of \acp{VSA}, there are systematical limitations to the amount of information that can be encoded in such a vector representation.
These limitations are strongly connected to the chosen dimension of the underlying vector space and are a feature of such modeling architectures for being able to model limitations of cognitive functions of living beings, who are also not able to store unlimited amounts of information.
Considering human subjects for instance, the capacity to process and store information or concepts in short-term memory as well as other cognitive tasks is subject to numerical restrictions \cite{Miller1956}.
Hence, numerical limitations of cognitive architectures like \acp{VSA} are one way of modeling the numerical restrictions to cognition observed in human subjects.
In our context of interest, i.e., automated driving \cite{Mirus2019b}, however, we need to analyze these restrictions imposed by \acp{VSA} in general and the \ac{SPA} in particular to provide upper borders regarding the amount of information that can be stored in our vector representation.

\subsection{Contribution}%
\label{subsec:contribution}

In this paper, we analyze the limits regarding information capacity of distributed representations with the goal of finding bounds for, e.g., the number of concepts that can effectively be stored in a single vector before the accumulation of noise makes it impossible to retrieve the original individual vectors.
Therefore, our contribution is a two-stage analysis: First, we analyze the amount of information that can effectively be stored in a single vector through superposition (i.e., addition) of several concept vectors.
A similar but slightly different experiment has been conducted in \cite{Wahle2012}: the atomic vocabulary vectors, referred to as elemental vectors in \cite{Wahle2012}, are sparse in the sense, that they mostly contain \num{0} elements, and the superposed vectors are normalized after adding them.
Furthermore, \cite{Wahle2012} only compares the similarity between the superposition and the original vector with the similarity between the original vector and the most recently added random vector as baseline for the expected similarity between randomly chosen vectors.
In contrast, we calculate the similarity between the superposition vector and $n$ other random vectors for reference.

Secondly, we analyze the information capacity of vector representations involving the convolutive vector power for encapsulating spatial information.
Given our scene representation proposed in \cite{Mirus2019b} (cf. Fig.~\ref{fig:spa_power_scene} and Sec.~\ref{sec:materials_and_methods}), we are primarily interested in representing two-dimensional values in vectors, which is why we focus our analysis of the convolutive power encoding scheme on this case.
In our analysis, we show that the information capacity is tightly linked to the dimension of the underlying vector space and furthermore, we give upper bounds for the capacity of superposition and convolutive power representations for three different vector dimensionalities.

\section{Materials and Methods}%
\label{sec:materials_and_methods}

\subsection{Convolutive vector-power}%
\label{subsec:convolutive_vector_power}

The \ac{SPA} \cite{Eliasmith2013} is based on Plate's \acp{HRR} \cite{Plate1994}, which is one special case of a \acl{VSA} \cite{Gayler2003}.
Here, atomic vectors are picked from the real-valued unit sphere, the dot product serves as a measure of similarity, which we denote by $\phi$, and the algebraic operations are component-wise vector addition $\oplus$ and circular convolution $\otimes$.
In this work, we make use of the fact that for any two vectors $\mathbf{v},\mathbf{w}$, we can write
\begin{equation}
  \mathbf{v} \otimes \mathbf{w} = \acs{IDFT} \left(\acs{DFT}(\mathbf{v}) \odot \acs{DFT}(\mathbf{w})\right),
  \label{eq:conv_dft}
\end{equation}
where $\odot$ denotes element-wise multiplication, \acs{DFT} and \acs{IDFT} denote the \acl{DFT} and \acl{IDFT} respectively.
Using Eq.~\eqref{eq:conv_dft}, we define the \emph{convolutive power} of a vector $\mathbf{v}$ by an exponent $p \in \mathbb{R}$ as
\begin{equation}
  \mathbf{v}^{p} := \Re\left(\acs{IDFT} \left(\left(\acs{DFT}_{j}\left(\mathbf{v}\right)^{p}\right)_{j=0}^{D-1}\right)\right),
  \label{eq:conv_power}
\end{equation}
where $\Re$ denotes the real part of a complex number.
Furthermore, we call a vector $\mathbf{u}$ \emph{unitary}, if $\norm{\mathbf{v}} = \norm{\mathbf{v} \otimes \mathbf{u}}$ for any other $\mathbf{v}$ (see \cite[Sec. 3.6.3 and 3.6.5]{Plate1994} for more details on the convolutive power and unitary vectors).

Finally, we consider any two vectors similar, if their similarity is significantly higher than what we would expect from two randomly chosen vectors.
For growing dimension $D$, the cosine similarity follows approximately a normal distribution $\mathcal{N}_{\mu, \sigma}$, with $\mu=0$ and $\sigma=\tfrac{1}{\sqrt{D}}$ \cite{Widdows2014}.
Using the three-sigma-rule, we denote $\epsilon_{weak} = \tfrac{2}{\sqrt{D}}$ as \emph{weak similarity threshold} and $\epsilon_{strong} = \tfrac{3}{\sqrt{D}}$ as \emph{strong similarity threshold}.

\subsection{Vector representation}%
\label{subsec:vector_representation}

\begin{figure*}[t]
    \centering
    \subfloat[\label{subfig:spa_power_representation_one_item}Convolutive power encoding for one two-dimensional numerical entity]{%
        \includegraphics[width=0.99\linewidth]{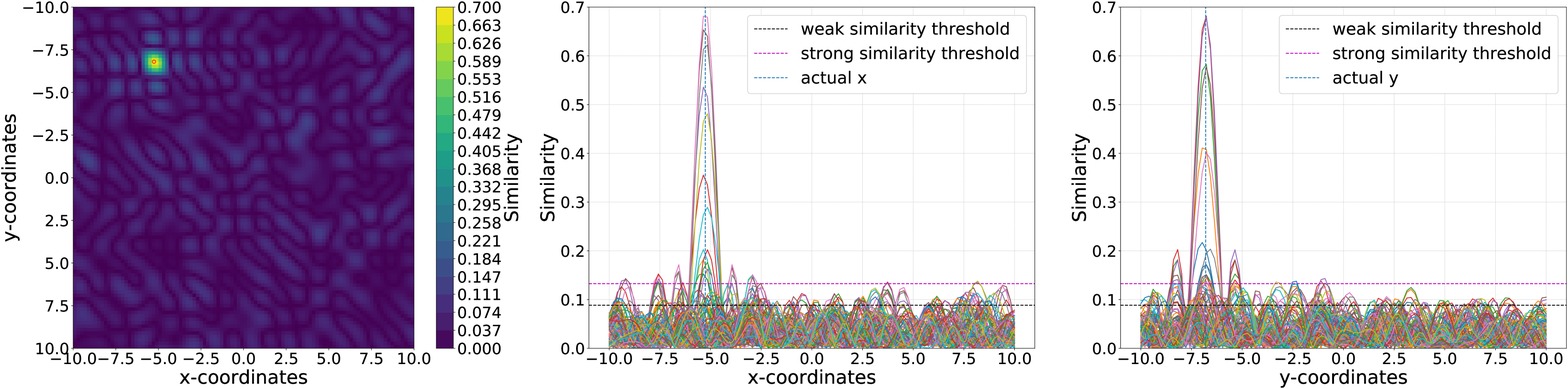}
    }\\
    \subfloat[\label{subfig:spa_power_representation_two_items}Convolutive power encoding for two two-dimensional numerical entities]{%
        \includegraphics[width=0.99\linewidth]{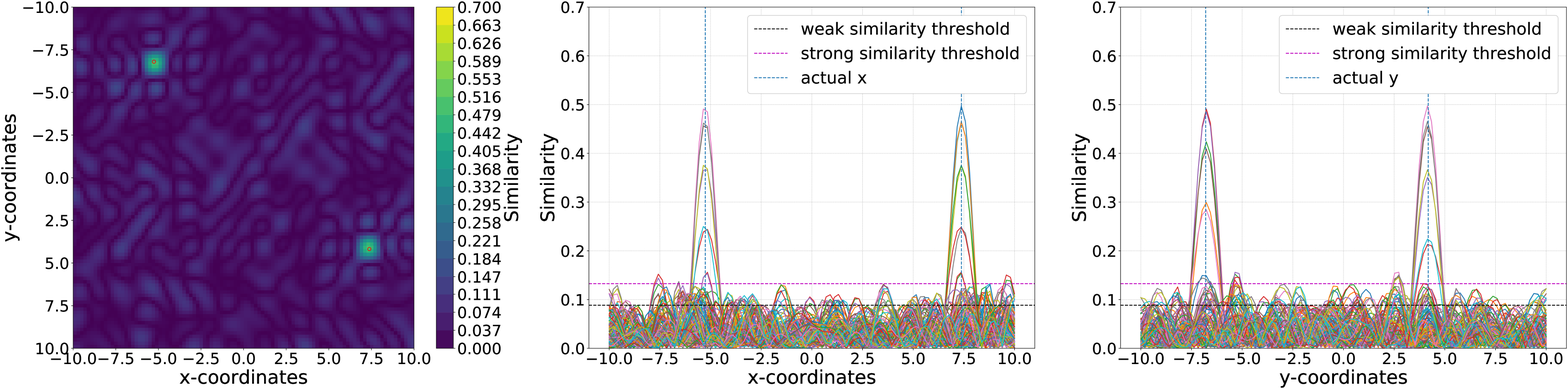}
    }
    \caption{
        Visualization of the convolutive power encoding scheme for \num{512}-dimensional representation vectors depicting the similarity between the representation vector and auxiliary comparison vectors created from a sequence of discrete values.
        The left plot in both rows shows a two-dimensional grid of the similarities, while the middle and right plot show the individual entities respectively.
        The red circles in the left plot and the dashed blue lines in the middle and right plots indicate the
    actual encoded values.}
    \label{fig:spa_power_encoding}
\end{figure*}

We are primarily interested in representing two-dimensional values in vectors, which is why we focus our analysis of the convolutive power encoding scheme on this case.
Hence, we encode two numerical values $x,y$, i.e., a two-dimensional entity, by generating two random, unitary vectors $ \mathbf{X}, \mathbf{Y} $ representing the corresponding units and applying Equation~\eqref{eq:conv_power} 
\begin{equation}
\label{eq:conv_power_2d}
\mathbf{V} = \mathbf{X}^{x} \otimes \mathbf{Y}^{y}.
\end{equation}
To encode a sequence $ \left(x_{i}, y_{i}\right) $ for $i=1, \ldots, n$ of two-dimensional numerical values all sharing the same units, we simply sum up a their individual encoding vectors generated via Equation~\eqref{eq:conv_power_2d}, which leads to
\begin{equation}
\label{eq:conv_power_sum}
\mathbf{V} = \sum\limits_{i=1}^{n} \mathbf{X}^{x_i} \otimes \mathbf{Y}^{y_i}.
\end{equation}
Figure~\ref{fig:spa_power_encoding} visualizes vectors encoding one (cf.\ Fig.~\ref{subfig:spa_power_representation_one_item} and Equation~\eqref{eq:conv_power_2d}) and two numerical entities (cf.\ Fig.~\ref{subfig:spa_power_representation_two_items} and Equation~\eqref{eq:conv_power_sum}) given by two units within one vector.
To generate the similarities shown in Fig.~\ref{fig:spa_power_encoding}, we calculate the dot product between the vectors actually representing the encoded values and vectors $ \tilde{\mathbf{V}}_{i} = \mathbf{X}^{\tilde{x}_{i}} \otimes \mathbf{Y}^{\tilde{y}_i} $ encoding a sequence of discrete sample values $ \left( \tilde{x}_{i}, \tilde{y}_{i} \right)$ for $i=1, \ldots, m$.
The left plot in each row of Fig.~\ref{fig:spa_power_encoding} depicts the similarities as heat map over a two-dimensional grid. 
The middle and right plots in Fig.~\ref{fig:spa_power_encoding} visualize the similarities of each unit, which is similar to plotting the heat map in three dimensions as ridges and slicing them through one of the ground axes.
In both rows, we observe high similarity peaks way above both similarity thresholds at the actual encoded values and significantly lower similarity values everywhere else.
However, comparing the similarities at the positions of the encoded values, we observe a drop of similarity values from roughly \num{0.7} to \num{0.5} when encoding two two-dimensional numerical values instead of only one.
This gives a graphic explanation that there is a limit of how many values can effectively be encoded in such a representation before the actual values can not be properly recovered anymore.
Such limitations regarding the number of concepts that can be represented in one vector depending on its dimension are a recurrent theme in the field of \acp{VSA} and the subject of our analysis in the following section.

\section{Experiments}%
\label{sec:experiments}

\begin{figure*}[t]
	\centering
	\includegraphics[width=0.95\textwidth]{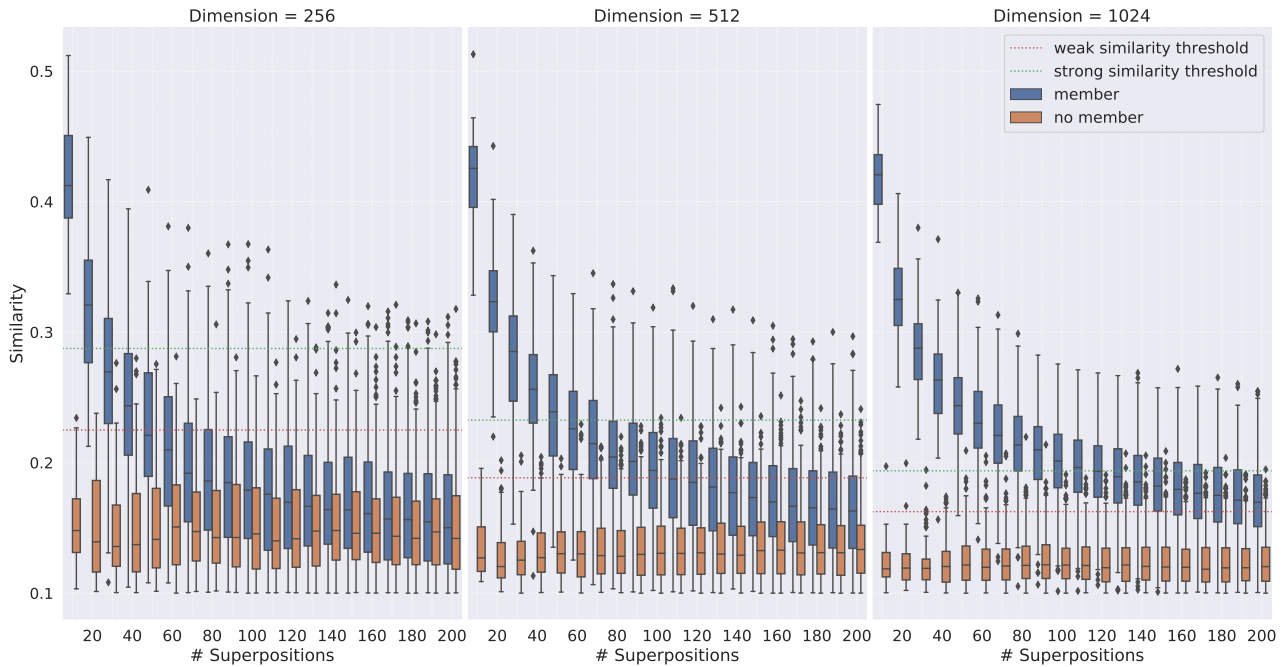}
	\caption{Visualization of the \ac{SPA}'s superposition capacity for vector dimensions \num{256}, \num{512} and \num{1024}.
    The blue boxes indicate the similarity between the superposition vector and its summands, the orange boxes illustrate the similarity between the superposition vector and other randomly generated vectors.
    The dotted lines visualize the similarity threshold based on the vector dimensionality for reference.}
	\label{fig:spa_superposition_capacity}
\end{figure*}

In this section, we conduct our analysis regarding the capacity of distributed representations for simple superposition in Sec.~\ref{subsec:superposition_capacity} and representations employing the convolutive power in Sec.~\ref{subsec:capacity_of_structured_representations_involving_convolutive_powers}.
The code for reproducing our analysis, results and figures can be found online \footnote{\url{https://github.com/fmirus/spa_capacity_analysis}}.

\subsection{Superposition capacity}%
\label{subsec:superposition_capacity}
First, we evaluate the capacity of superposition, i.e., the addition operation.
Superposition is used to store and combine several concept vectors $ \mathbf{v}_i$ for $i=0, \ldots, n$ in an unordered set
\begin{equation}
\label{eq:superposition}
\mathbf{s} = \sum\limits_{i=0}^{n} \mathbf{v}_{i}.
\end{equation}
We can determine if a vector of interest $\mathbf{w}$ belongs to that ordered set by calculating the similarity $\phi\left( \mathbf{s}, \mathbf{w}\right)$ between the superposition vector and the vector of interest.
For sufficiently high-dimensional vectors, the similarity $\phi\left( \mathbf{s}, \mathbf{w}\right)$ will be close to \num{0} in case the vector $ \mathbf{w}$ is not part of the sum.
However, the more vectors we add to the superposition vector $ \mathbf{s}$, the more noise accumulates in the representation and thus decreases the similarity between the superposition vector $ \mathbf{s}$ and its individual ingredients $ \mathbf{v}_i$.
In order to analyze how many vectors can be added together by superposition before individual vectors become irretrievable, we conducted the following experiment: assuming we want to add $n$ vectors $ \mathbf{v}_i$ for $i=1, \ldots, n$ into a superposition vector $ \mathbf{s}$ as in Equation~\eqref{eq:superposition}, we randomly generate a vocabulary of $2n$ vectors $ \mathbf{v}_i$ for $i=1, \ldots, 2n$ and sum up the first $n$ members to create our superposition vector $ \mathbf{s}$.
Then we calculate the cosine similarity $\phi\left(\mathbf{s}, \mathbf{v}_i\right)$ between the superposition vector $ \mathbf{s}$ and every vector $ \mathbf{v}_{i}$ for $i=1,\ldots, 2n$ in the vocabulary.

Figure~\ref{fig:spa_superposition_capacity} shows the result of our experiment for \num{3} random vocabularies per superposition length containing vectors of dimension \num{256}, \num{512} and \num{1024}.
The blue boxes in each figure illustrate the similarity between the superposition vector $ \mathbf{s}$ and each of the individual vectors $ \mathbf{v}_i$ for $i=1, \ldots, n$ it contains, i.e., the members of the unordered superposition set.
The orange boxes depict the similarity between $ \mathbf{s}$ and the other vocabulary vectors $ \mathbf{v}_i$ for $i=n+1, \ldots, 2n$ it does not contain, i.e., the non-members.
The dotted red and green lines indicate the \ac{SPA}'s weak and strong similarity threshold depending on the dimension of the vector space.
Considering the weak similarity threshold $\epsilon_{weak} = \tfrac{2}{\sqrt{D}}$, we observe that for a vector dimension of \num{256} the \ac{SPA} allows roughly \num{50} items to be stored in a superposition vector.
For higher vector dimensions \num{512} and \num{1024}, the number of items that can be superposed increases to roughly \num{100} and \num{200} respectively.
Considering the strong similarity threshold $\epsilon_{strong} = \tfrac{3}{\sqrt{D}}$, the upper borders for the number of items being stored in a superposition vector are slightly more conservative with \num{25}, \num{50} and \num{100} for vector space dimensions of \num{256}, \num{512} and \num{1024} respectively.
We also observe in our experiments that the similarity between the superposition vector and non-member random vectors is consistently below the weak similarity threshold $\epsilon_{weak}$ for the majority of the samples.
However, once the similarity between the superposition vector and its members drops below either of the similarity thresholds for the majority of the samples, we can not distinguish between members and non-members with a sufficiently high probability.
For \num{256} dimensional vectors for instance, we even observe that the members and non-members become nearly indistinguishably when adding more than \num{80} vectors.
Thus, we have to choose rather conservative bounds for the number of items to be encoded in a superposition vector.

\begin{figure*}[t]
    \centering
    \includegraphics[width=0.95\linewidth]{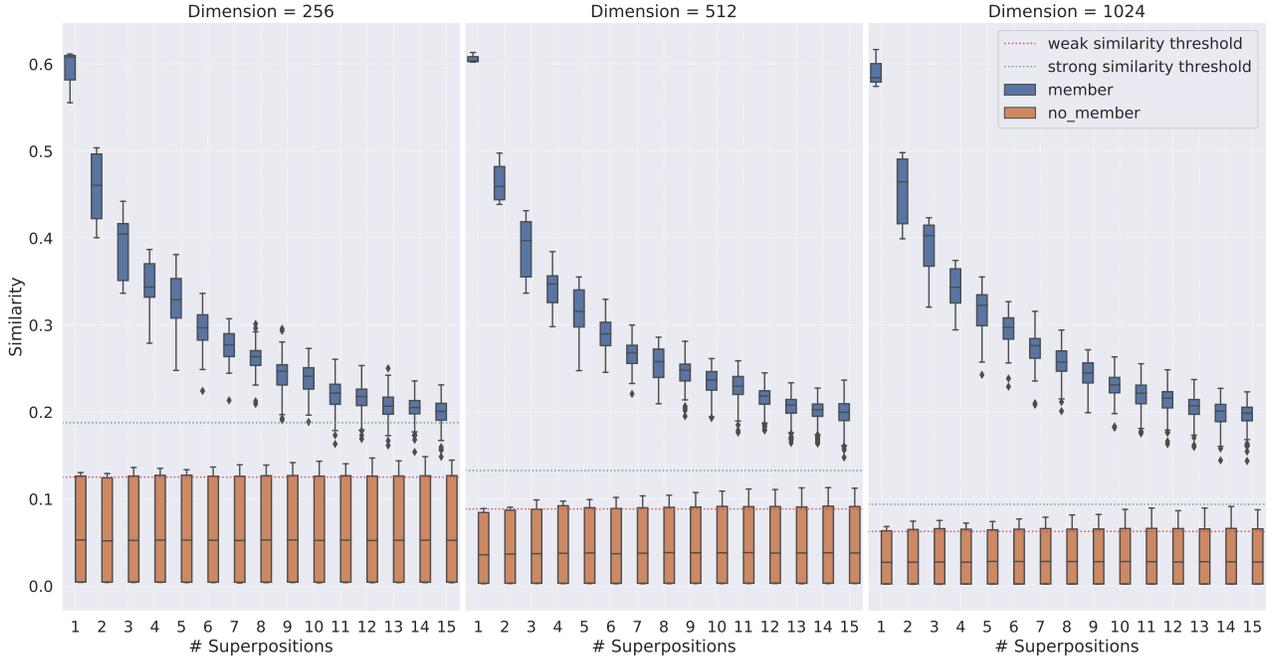}
    \caption{Capacity analysis for the superposition of vectors encoding spatial positions using the convolutive vector-power for varying vector dimensions.}
    \label{fig:spa_power_capacity}
\end{figure*}

\subsection{Capacity of structured representations involving convolutive powers}%
\label{subsec:capacity_of_structured_representations_involving_convolutive_powers}

In the previous section, we have analyzed the \ac{SPA}'s capacity regarding the number of items that can be stored in an unordered set using superposition.
For encoding more complex information, such as driving situations \cite{Mirus2019b}, in a semantic vector substrate, we employ more complex representation than superposition of single items alone.
In this section, we analyze the capacity of structured vector representations involving the convolutive vector-power (see Eq.~\eqref{eq:conv_power}).
Figure~\ref{fig:spa_power_encoding} illustrates that we can unbind positions by querying the representation vector with sample vectors encoding discrete position examples, but also that encoding several entities of the same type in one position vector as in Fig.~\ref{subfig:spa_power_representation_two_items} yields lower similarities of the true positive positions compared to the encoding of only one item as in Fig.~\ref{subfig:spa_power_representation_one_item}.
Hence, our capacity analysis has to cover the amount of objects encoded in one vector, but also the number of items per object class.

\begin{figure*}[t]
    \centering
    \includegraphics[width=0.95\linewidth]{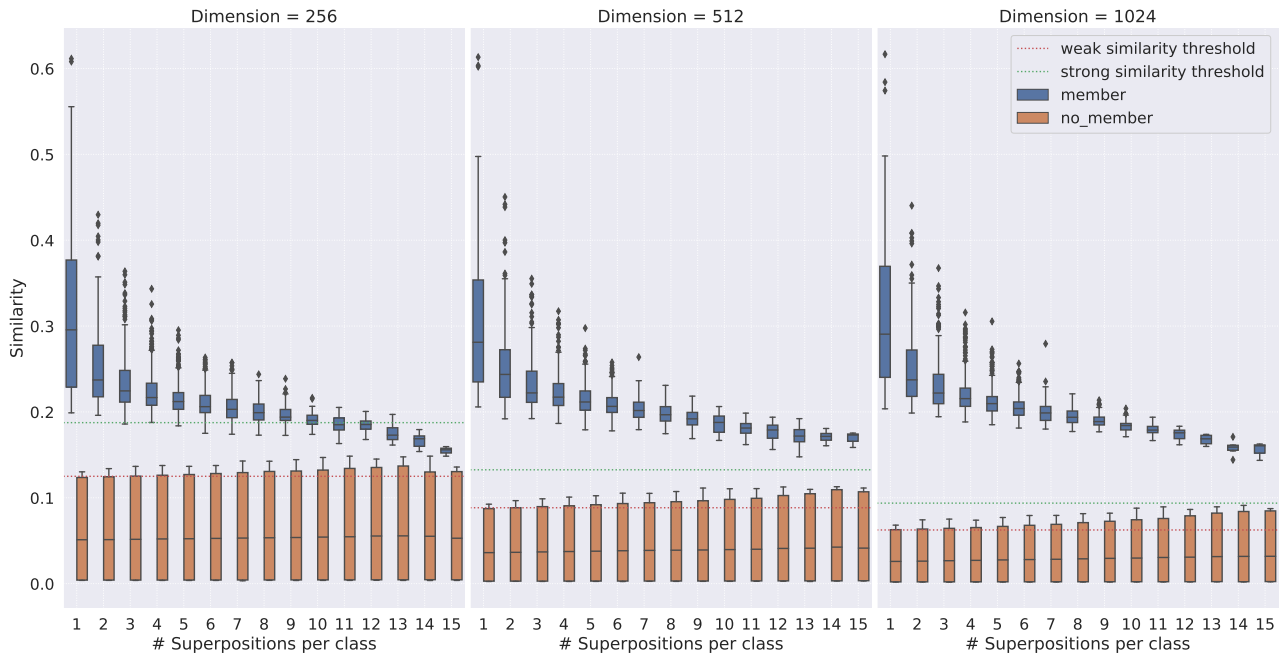}
    \caption{Capacity analysis for the superposition of vectors encoding spatial positions using the convolutive vector-power for varying vector dimensions.
        In contrast to Fig.~\ref{fig:spa_power_capacity}, this figure illustrates the similarity for vectors containing spatial information for several objects of the same class.
    }
    \label{fig:spa_power_capacity_superpositions_per_class}
\end{figure*}

Therefore, we conduct the following experiment: assuming we want to encode $n$ spatial entities, i.e., objects $o_{i}$ with two-dimensional location information $ \left(x_{i}, y_{i}\right)$ for $i=1, \ldots, n$ as shown in Fig.~\ref{fig:spa_power_encoding} for $n=1$ (Fig.~\ref{subfig:spa_power_representation_one_item}) and $n=2$ (Fig.~\ref{subfig:spa_power_representation_two_items}), into a single representation vector $ \mathbf{s}$, we generate a vocabulary of random vectors $
\mathbf{v}_{i}$ for $i=1, \ldots, n$ encoding object class labels and random unitary vectors $ \mathbf{X}, \mathbf{Y}$ to encode the units of the spatial information.
In contrast to the experiment in~\ref{subsec:superposition_capacity}, where we simply summed up a certain number of random vectors, we are interested in a more specific analysis, since there are several possibilities to distribute the positional values $ \left(x_{i}, y_{i}\right)$ over the available object class vectors $ \mathbf{v}_{i}$.
For instance, for a total number of two superpositions, i.e., $n=2$, there are two possibilities to generate our representation vector, namely
\begin{align}
    \mathbf{s}_{1} &= \mathbf{v}_{1} \otimes \mathbf{X}^{x_{1}} \otimes \mathbf{Y}^{y_{1}} + \mathbf{v}_{1} \otimes \mathbf{X}^{x_{2}} \otimes \mathbf{Y}^{y_{2}}, \label{eq:spa_power_exp_v1} \\
    \mathbf{s}_{2} &= \mathbf{v}_{1} \otimes \mathbf{X}^{x_{1}} \otimes \mathbf{Y}^{y_{1}} + \mathbf{v}_{2} \otimes \mathbf{X}^{x_{2}} \otimes \mathbf{Y}^{y_{2}}. \label{eq:spa_power_exp_v2} 
\end{align}
The vector $ \mathbf{s}_{1}$ in Equation~\eqref{eq:spa_power_exp_v1} encodes two objects of the same type, while the vector $ \mathbf{s}_{2}$ encodes occurrences of two different object types at the given locations.
As we are working with random vectors in this experiment, we can, without loss of generality, skip the vector encoding two objects of type represented by the vector $ \mathbf{v}_{2}$, which would yield a result equivalent to Equation~\eqref{eq:spa_power_exp_v1}.
More generally, we are interested in all sets
\begin{equation}
\label{eq:sum_combinations}
C_{m,j} = \left\{0 < k_{1}, \ldots, k_{m} \leq n \quad | \quad m \leq n \textrm{ and } \sum\limits_{i=1}^{m} k_{i} = n \right\}
\end{equation}
of natural numbers $k_{i}$ summing up to the total number of objects $n$ ignoring permutations of the $k_{i}$. 
We index the sets with $j$, since there potentially exist several possibilities to decompose $n$ into sums of $m$ natural numbers.

In our experiments, for each number $n$ of total objects to be encoded in the vector representation, we calculate all possible sets $C_{m,j}$ (ignoring permutations) and generate random position values $ \left(x_{i}, y_{i}\right)$ for $i=1, \ldots, n$ and a random vocabulary as described above.
For each set $C_{m,j}$, we generate a representation vector 
\begin{equation}
\label{eq:spa_power_exp_v_gen}
\mathbf{s}_{m,j} = \sum\limits_{i=1}^{m} \sum\limits_{l=1}^{k_{i}} \mathbf{v}_{i} \otimes \mathbf{X}^{x_{i}} \otimes \mathbf{Y}^{y_{i}}
\end{equation}
as well as query vectors $ {\mathbf{P}}_{i} = \mathbf{X}^{\tilde{x}_{i}} \otimes \mathbf{Y}^{\tilde{y}_i} $ encoding a sequence of discrete sample values $ \left( \tilde{x}_{i}, \tilde{y}_{i} \right)$ for $i=1, \ldots, M$ evenly distributed over the length of the positional encoding grid.
In other words, Equation~\eqref{eq:spa_power_exp_v_gen} states, that each class label $ \mathbf{v}_{i}$ appears $k_{i}$ times yielding a sum of $n$ objects.
We query the representation vector for the position of each class by binding it to the pseudo-inverse (see also \cite{Plate1994, Eliasmith2013}) element $ \bar{ \mathbf{v}}_{i}$ for each class label vector, i.e., 
\begin{equation}
\label{eq:spa_power_query}
\mathbf{s}_{m,j} \otimes \bar{ \mathbf{v}}_{i} \approx \sum\limits_{l=1}^{k_{i}} \mathbf{X}^{{x}_{i}} \otimes \mathbf{Y}^{{y}_i}, 
\end{equation}
and calculate the similarity with the discretized position vectors $ \mathbf{P}_{k}$ to get
\begin{equation}
\label{eq:text}
s_{i,k} = \left|\phi \left( \mathbf{s}_{m,j} \otimes \bar{ \mathbf{v}}_{i}, \mathbf{P}_{k}\right) \right|.
\end{equation}
For samples close to the originally encoded positions, i.e., $\left| x_{i} - \tilde{x}_{i}\right| < \epsilon$ and $\left| y_{i} - \tilde{y}_{i}\right| < \epsilon$ for a certain threshold $\epsilon$ (here, we use $\epsilon=0.4$), we label $s_{i,k}$ as positive similarity denoting a member of the representation vector.
Otherwise, we consider the similarity $s_{i,k}$ at position $\left( \tilde{x}_{i}, \tilde{y}_{i} \right)$ not a member of the representation vector $ \mathbf{s}_{m,j}$.

Figure~\ref{fig:spa_power_capacity} shows the results of this capacity analysis regarding the total number of superposed objects within the representation vector for varying vector dimensions.
In Sec.~\ref{subsec:superposition_capacity}, we have already analyzed the \ac{SPA}'s capacity regarding the number of items that can be stored in an unordered set using superposition.
Similar to Fig.~\ref{fig:spa_power_encoding}, we observe that the similarity of the non-members is in the order of magnitude of the similarity thresholds while the similarity for the member position decreases with a growing number of spatial items encoded in the vector.
However, for encoding more complex information, like automotive scenes \cite{Mirus2019b} in a semantic vector substrate, we employ more complex representation than superposition of single items alone.

Therefore, Fig.~\ref{fig:spa_power_capacity_superpositions_per_class} shows a different evaluation of the same data showing the number of addition operations per class on the $x$-axis.
In contrast to Fig.~\ref{fig:spa_power_capacity}, Fig.~\ref{fig:spa_power_capacity_superpositions_per_class} illustrates the similarity for vectors containing spatial information for several objects of the same class.
That is, Fig.~\ref{fig:spa_power_capacity_superpositions_per_class} illustrates the similarity of vectors containing a specific number $k$ of superpositions per class on its $x$-axis independent of the total number of superpositions.
We observe, that not only the similarity of the members decreases with growing number of superpositions per class, but the similarity of the non-member increases beyond the weak similarity threshold.
Similar to the simple superposition capacity analysis, we consider the point in the plots where the member similarities fall below the strong similarity threshold the upper border for the maximal number of spatial objects per class to be encoded in this representational substrate.
For instance, this upper bound for \num{256} dimensional vectors is \num{10} superpositions per class, which is roughly half of the upper bound for the number of simple superpositions.
For higher-dimensional vectors (here \num{512} and \num{1024}), these limits are beyond the evaluated number of superpositions per class.
Therefore, using at least \num{512} dimensional vectors for automotive scene representation yields a sufficiently high information capacity.
On the other hand, we expect upper bounds for the higher dimensions similar to \num{256}-dimensional vectors, i.e., roughly half the number of superpositions as stated in Sec.~\ref{subsec:superposition_capacity}.

\section{Conclusion}%
\label{sec:conclusion}

In this paper, we analyzed the capacity of structured vector representations in \acp{VSA} based on simple superposition and superposition combined with the convolutive power encoding of spatial information.
We provided a more detailed analysis of the superposition capacity compared to those available in the literature, e.g.\ in \cite{Wahle2012}. 
Furthermore, we evaluated the capacity of structured representations involving convolutive vector powers to encode spatial information, which, to the best of our knowledge, is the first of its kind. 
Thereby, we found upper bounds for the amount of information that can effectively be encoded in such representations depending on the dimension of the underlying vector space.
These bounds have to considered in future work regarding distributed representations, for instance, of automotive scenes like in \cite{Mirus2019b} to evaluate if the amount of information to be encoded in the vector representation is compliant with these bounds.
This will allow a conclusive assessment of the limits of structured vector representations in general, whereas our particular focus is on automotive context.

\IEEEtriggeratref{12}
\bibliographystyle{IEEEtran}
\bibliography{literature}

\end{document}